\documentclass{article}
\usepackage{neurips_2022}[preprint=true]

\usepackage{tikz}

\usepackage{natbib}
\setcitestyle{square, comma, numbers,sort&compress, super}
\usepackage[inline]{enumitem} 
\usepackage{amssymb}
\usepackage{graphicx}
\usepackage{color}
\usepackage{xcolor}
\usepackage[caption=true,font=footnotesize]{subfig}
\usepackage[linesnumbered,ruled,vlined]{algorithm2e}
\usepackage{circledsteps}

\usepackage{mathtools} 

\DeclareMathOperator*{\arctanh}{arctanh}

\usepackage{dsfont}

\usepackage{amstext} 
\usepackage{array}   
\newcolumntype{L}{>{$}c<{$}} 

\usepackage{booktabs}

\newcommand{\sysname}{HideNseek}

\DeclareMathOperator{\sign}{sign}

\usepackage{wrapfig}

\begin{document}

\title{
    \sysname: Federated Lottery Ticket via \\
    Server-side Pruning and Sign Supermask
}


\maketitle

\begin{abstract}
 
Federated learning alleviates the privacy risk in distributed learning by transmitting only the local model updates to the central server. However, it faces challenges including statistical heterogeneity of clients' datasets and resource constraints of client devices, which severely impact the training performance and user experience. Prior works have tackled these challenges by combining personalization with model compression schemes including quantization and pruning. However, the pruning is data-dependent and thus must be done on the client side which requires considerable computation cost. Moreover, the pruning normally trains a binary supermask $\in \{0, 1\}$ which significantly limits the model capacity yet with no computation benefit. Consequently, the training requires high computation cost and a long time to converge while the model performance does not pay off. In this work, we propose \sysname\ which employs one-shot data-agnostic pruning at initialization to get a subnetwork based on weights' synaptic saliency. Each client then optimizes a sign supermask $\in \{-1, +1\}$ multiplied by the unpruned weights to allow faster convergence with the same compression rates as state-of-the-art. Empirical results from three datasets demonstrate that compared to state-of-the-art, \sysname\ improves inferences accuracies by up to 40.6\% while reducing the communication cost and training time by up to 39.7\% and 46.8\% respectively.

\end{abstract}

\section{Introduction}

Federated learning~\citep{mcmahan} improves privacy protection by decoupling the need for a central data repository from learning distributed datasets. Each client uploads its local model trained with local datasets to a central server that maintains a global model.
The server updates the global model via aggregating the clients' updates and sends the new model to the clients. These processes are iterated until the model converges or the predefined cycle ends. 

While improving privacy protection, this paradigm also faces challenges such as statistical heterogeneity, which refers to the non-IID distribution of the data among clients, impacting the global model convergence if data from different clients are too diverged. Some works address statistical heterogeneity by 
introducing various learning techniques for personalization~\citep{fallah2020personalized, smith2017federated, ensemble_distill, attn_distill,data_free}. 

Another major challenge faced by federated learning is the resource constraints, which refer to the limited computation capacity and transmission bandwidth of the client devices. The limited resource restricts the model size that can be trained in the client devices and transmitted timely to the server for aggregation. Several recent works have surged to address this challenge by 
adapting training procedures \citep{FedProx, reisizadeh2020fedpaq} and employing model compression schemes~\citep{heterofl, fjord, federated_dropout}.

Lately, FedMask~\citep{li2021fedmask} tackles both statistical heterogeneity and resource constraints by employing a ``masking is training'' philosophy based on the lottery ticket hypothesis~(LTH)~\citep{LTH}. FedMask prunes the local model to comply with the resource constraints. Then, the client learns a sparse local binary supermask for personalization to alleviate the statistical heterogeneity across clients. 

However, exiting methods face several challenges. First, current pruning methods in federated setup are mostly data-dependent and thus have to be performed on the client-side. Hence the client devices with limited resources cannot avoid the considerable computation cost. 
Second, such pruning methods commonly employ the binary supermask which essentially is unstructured pruning that brings no computational advantage yet limits the model capacity. 
Consequently, the model performance is limited, yet requires high computation cost and a long convergence period.

Therefore, in this paper, we propose \sysname, a statistical heterogeneity-aware federated learning framework that provides computation and communication efficiency powered by sign supermask \citep{zhou2019deconstructing}. Specifically, we make the following contributions.
\begin{itemize}
    \item \sysname\ proposes a federated version of LTH with sign optimization for the first time. Compared with the commonly used binary supermask, our approach provides higher accuracy and faster convergence. 
    \item \sysname\ performs server-side one-shot pruning at initialization by employing an iterative data-agnostic approach based on synaptic saliency of the weights' signs. As such, \sysname\ greatly alleviates the computation burden and communication cost for the clients with limited capacities.
    \item Empirical results on varied datasets demonstrate that \sysname\ outperforms the state-of-the-art methods in inference accuracy by up to 40.6\% while reducing the communication cost and training time by up to 39.7\% and 46.8\% respectively.
\end{itemize}

\section{Background \& Objective Formulation}
In this section, we begin with the background on federated learning and LTH and formulate the objective of data-agnostic pruning for federated learning.  
\subsection{Federated Learning}
In federated learning, a deep neural network must be learned in a distributed fashion. This is achieved by a central server aggregating copies of the network weights learned among clients on their local datasets. The learning objective is to find weights $w$ that minimize the empirical loss across clients 
\begin{equation}
    \min_{w} F(w) = \sum_{k=1}^{K} \frac{n_k}{n} \mathcal{L}[f(x_k;w), y_k]
\end{equation}
where the network $f(\cdot;w)$ is a composite function of layers parametrized by vectorized weights $w \in \mathbb{R}^d$ and $\mathcal{L}$ is the empirical loss function which measures the ability to approximate the function generating the local dataset $(x_k, y_k)$ of client $k$. $n_k$ is the number of local samples and $n = \sum_{k}n_k$ is the total number of samples across all the $K$ clients. 

A prior work \citep{mcmahan} provides the widely used FedAvg to solve this objective via distributed SGD. In each communication round $t$, the central server selects a subset of $c \ll K$ clients and sends them a copy of the global weights $w^t$. The clients modify their local copy of the weights $w_k^t$ by minimizing the empirical loss on local datasets to obtain $w_k^{t+1}$ and transmit them back to the server. The server updates the global weights by simply averaging the clients' weights
\begin{equation}
    w^{t+1} = \frac{1}{c} \sum_{k=1}^c w_k^{t+1}
\end{equation}
Several problems arise in a real-world deployment. Firstly, the clients incur communication costs due to weight transmission and computation costs due to local weight optimization. Secondly, the local datasets are not independent and identically distributed (non-IID) so the global weights must generalize well among clients. Some works have modified their learning objective(s) to address these problems (see Section~\ref{sec:related}). 

\subsection{Lottery Ticket Hypothesis}
LTH \citep{LTH} is a nascent avenue in machine learning that can tackle the aforementioned issues in federated learning. It states that a randomly initialized neural network contains a subnetwork called a winning ticket when trained in isolation, performs as well as the original network. 
%
A more ambitious extension comes from 
\citep{hidden_nets} which states that a sufficiently overparametrized network contains a winning ticket at a randomly initialized state. Furthermore, this winning ticket can be determined via freezing the weights at random initialization $w^0$ and pruning a subset of these weights to find a sparse subnetwork. To improve the learning performance, 
\citep{peekaboo} suggest applying a transform function $U \in \mathcal{U}$ to the weights to further minimize the empirical loss. The learning objective is thus
\begin{equation} \label{eq:3}
    \min_{U \in \mathcal{U}, ||m||_{0}=S} \mathcal{L}[f(x; U(w^0 \odot m)), y]
\end{equation}
where $m \in \{0, 1\}^d$ is a binary supermask with sparsity level $S$ and the same dimensionality as the weights. The winning ticket is represented as the element-wise multiplication of the weights and the supermask, i.e., $w^0 \odot m$. However, optimizing Eq.~\eqref{eq:3} is computationally intractable due to the large dimensionality of weights and transformation space $\mathcal{U}$. Hence, they propose to decouple the optimization into two stages. The first stage is the pruning phase where a binary supermask must be found to sparsify the model by optimizing the following
\begin{equation}
    \hat{m} \in \min_{||m||_{0}=S} R(f(x; w^0 \odot m))
\end{equation}
where $R$ is a scoring function that measures the ability of the binary supermask $m$ to isolate the winning ticket from the model. The second stage is the training phase where a weight transformation is learned to minimize the empirical loss
\begin{equation}
    \hat{U} \in \min_{U \in \mathcal{U}} \mathcal{L}[f(x; U(w^0 \odot \hat{m})), y]
\end{equation}

\subsection{Objective Formulation}
In essence, LTH applies a ``masking is training'' philosophy where an optimal sparse subnetwork must be learned without modifying the weights. Given the communication and computation efficiency brought by this idea (discussed in subsequent sections), we propose to adopt this hypothesis for the federated setting. 
However, the weight transformation space $\mathcal{U}$ is very vast. Following \citep{peekaboo}, we confine the space to sign flipping transformation space $\mathcal{U}_s \in \mathcal{U}$ where a transformation $U(w,s) = w \odot s$ element-wise multiplication of the sign supermask $s \in \{-1, +1\}^d$ to the weights $w$. The updated learning objective of our work is 

\begin{equation}
    \label{eq:objective}
    \min_{m, s} F(m, s) = \sum_{k=1}^{K} \frac{n_k}{n} \mathcal{L}[f(x_k;w^0 \odot m \odot s), y_k]
\end{equation}
The subsequent sections elaborate on a federated learning algorithm to solve the above objective. 

\section{Methodology}
\subsection{\sysname}
\begin{figure*}
    \centering
    \includegraphics[width=\textwidth]{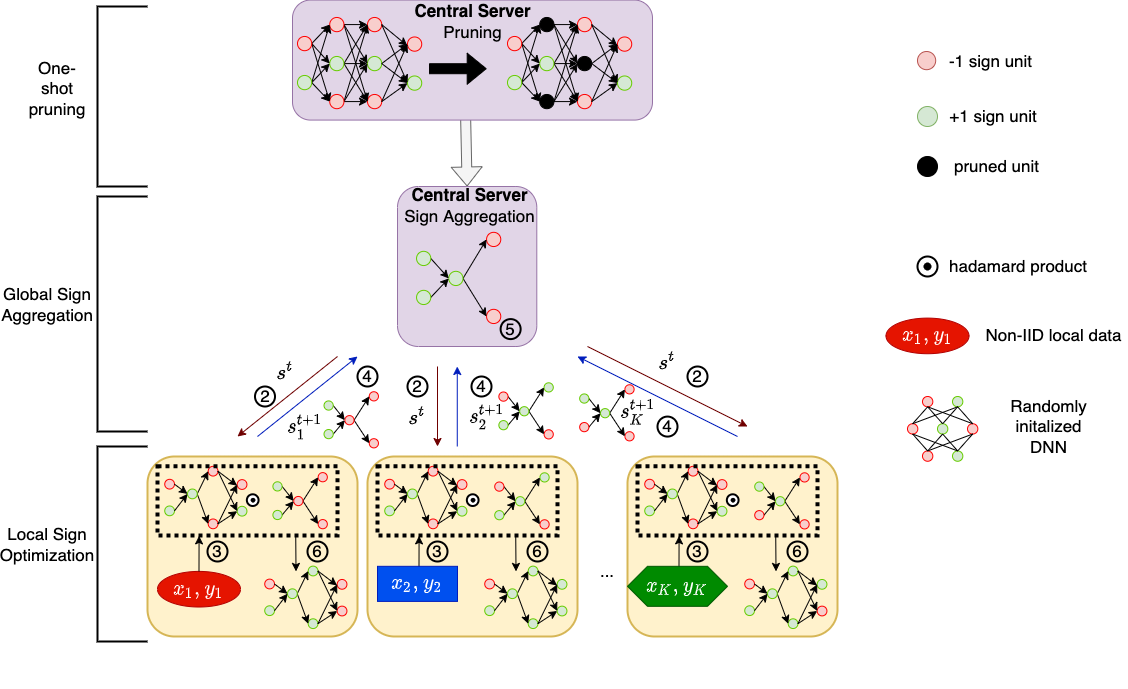}
    \caption{Overview of \sysname\ framework}
    \label{fig:overview}
\end{figure*}

In this work, we propose an efficient federated learning algorithm called \sysname\ by solving for federated adaptation of the LTH. Figure \ref{fig:overview} depicts an overview of the framework. As mentioned earlier, the learning process can be performed in two phases. 

In the first phase, the server first performs pruning-at-initialization to isolate the winning ticket (\Circled{1}). Since the server does not possess training data, a data-agnostic pruning method is applied. 
Following \citep{synflow}, we measure the score of the signs of weights via their synaptic saliency (see Section~\ref{sec:server_side_pruning}), 
and employ global structured pruning for hardware efficiency. 

In the second phase, an optimal weight transformation has to 
be learned to minimize the empirical loss in a federated manner. In each training step $t$, the server sends the global sign supermask $s^t$ to the selected clients (\Circled{2}) which initializes a local sign supermask $s^{t+1}_k$. The clients then freeze the model weights and optimize the local sign supermask by minimizing the empirical loss using Eq.~\eqref{eq:objective} (\Circled{3}). As such, a sign flipping transformation is learned and sent back to the server (\Circled{4}). The server then aggregates these local supermasks using Eq.~\eqref{eq:aggregation} (\Circled{5}).  

After the training phase, each client multiplies the aggregated sign supermask to its weights to get the final local model (\Circled{6}). Algorithm \ref{alg:hidenseek} (in Appendix \ref{sec:algorithm}) summarizes the processes with highlighted 
details elaborated in the following paragraphs. 

\subsection{Personalization}
\label{sec:personalization}

The output layer of the model cannot be simply optimized for its signs because the weight magnitudes must be scaled for stable training. We thus split our models into feature extractors which constitute all the hidden layers and classifier which is the output layer. The weights are frozen and the sign supermask is learned only for the feature extractor while the weights of the classifier are modifiable, as the optimization of 
the local data. Aligned 
with prior works \citep{data_free}, our work 
further reinforces 
personalization among clients. 

\subsection{Sign Flipping Transformation}
Optimizing the sign supermask is crucial to learning in our algorithm. As such, we handle certain preliminaries 
to achieve the optimization. As mentioned earlier, a model can be expressed as a composition of layers performing operations of vectorized weights. As an example, let us consider the fully connected layer. Note that the bias term has been omitted for brevity. A fully-connected layer $l$ can be expressed as $y^{[l]} = (w^{[l]} \odot s^{[l]}) \cdot x^{[l]}$, where $y^{[l]} \in \mathbb{R}^i$ is the output, $x^{[l]} \in \mathbb{R}^j$ is the input, and $w^{[l]} \in \mathbb{R}^{i \times j}$ is the weights. We thus handle the learning of 
a sign supermask $s^{[l]} \in \{-1, +1\}^{i \times j}$ with the same dimensionality as the weights. 

However, the traditional SGD optimization cannot be applied to the sign supermasks due to their discrete nature. Hence, we implement a straight-through estimator \citep{STE} with a real-valued sign supermask $\hat{s}^{[l]} \in \mathbb{R}^{i \times j}$. In the forward pass, $\hat{s}$ is quantized using the piecewise sign function
\begin{equation}
    s_{ij} = \sign({\hat{s}_{ij}}) = 
    \begin{cases}
        +1 & \hat{s}_{ij} \geq 0 \\
        -1 & \hat{s}_{ij} < 0
    \end{cases}
\end{equation}
where $s_{ij}$ is an element in the $i$-th row and $j$-th column of the sign supermask $s$. The gradients of $s$ in the backward pass are computed as $\nabla_{s} \mathcal{L} = (\nabla_y \mathcal{L} \cdot x^T) \odot w$. Given that the sign function is not differentiable, directly assigning the gradients from the quantized sign supermask to the real sign supermask (i.e., $\nabla_{\hat{s}} \mathcal{L} = \nabla_{s} \mathcal{L}$) would lead to large gradient variance \citep{binary_net}. Hence, we employ a hyperbolic tangent function, denoted as $\tanh(\cdot)$, for a continuous approximation of the sign function for the backward pass, $s_{ij} = \tanh(\hat{s}_{ij})$. This would allow us to compute the gradients of the real sign supermask $\hat{s}$ from the binary sign supermask $s$ as $\nabla_{\hat{s}} \mathcal{L} = \Psi \odot \nabla_s \mathcal{L}$, where $\Psi$ is the gradient matrix of the hyperbolic tangent function with values explicitly calculated as $\Psi_{ij} = (1 - \hat{s}_{ij}^2$). 

Since the clients transmit quantized sign supermasks to the server for communication efficiency, we employ the following sign aggregation scheme to obtain the real-valued sign supermask at the server
\begin{equation}
    \label{eq:aggregation}
    \hat{s} = \arctanh \Bigg( \sum_k \frac{n_k}{n} s_k \Bigg)
\end{equation}

\subsection{Server-side Pruning}
\label{sec:server_side_pruning}

Another crucial aspect in \sysname\ is the pruning phase where the winning ticket must be isolated from the network. Following \citep{li2021fedmask}, we employ one-shot pruning at initialization but perform it on the server-side to reduce the load on the clients. 
As the server does not contain any training data, we employ a data-agnostic iterative pruning approach \citep{synflow} where the prune scores are determined based on the weight's synaptic saliency. 
Given that the weights are frozen during training the sign supermask,
we measure the synaptic saliency of the sign of a given weight in a model with 
$L$ layers is as follows
\begin{equation}
    \label{eq:synaptic}
    R_{\text{SF}}(s^{[l]}_{ij}) = \left[\mathds{1}^\intercal\prod_{h=l+1}^{L}\left|s^{[h]} \odot w^{[h]} \right|\right]_i\left|s^{[l]}_{ij} \odot w^{[l]}_{ij}\right|\left[\prod_{h=1}^{l-1} \left|s^{[h]} \odot w^{[h]}\right|\mathds{1}\right]_j
\end{equation}
In essence, the synaptic saliency of a weight's sign is the product of all the weights multiplied by their respective sign supermasks that have the weight's sign within the path from the input to the output layer. To further promote hardware efficiency, we employ global structured pruning by scoring groups of weights by channels in convolutional layers and nodes in fully-connected layers. The prune score for the $i$-th channel or node in a layer as $\lVert w^{[l]}_i \odot \nabla_{s} R_{SF, i}^{[l]} \rVert_2$. Additionally, we keep the first few layers and only prune from the latter layers of the model with a prune rate $p_r$.





\subsection{State of the Art} 
FedMask \citep{li2021fedmask} is the closest in spirit to our work. However, its pruning method is data-dependent, and hence, have to 
be performed on client devices, resulting in 
increased 
computational load on the resource-constrained clients. 
Additionally, its weight transformation space is confined to the binary supermasking $\mathcal{U}_b \in \mathcal{U}$, where a local binary supermask $m_k \in \{0, 1\}$ is learned in the training stage in contrast to the binary sign supermask. Effectively, FedMask further performs unstructured pruning during the training stage with no communication or computation advantage. Contrarily, \sysname\ maintains all the weights after the pruning stage allowing for a greater model capacity. 



\section{Experimental Setup}
\label{sec:exp_setup}

\subsection{Datasets \& Models}

We evaluate \sysname\ on two applications, including image classification and human activity recognition, using the  EMNIST~\citep{caldas2018leaf} and  HAR~\citep{har_dataset} datasets, respectively. EMNIST is a handwritten character recognition task involving 28$\times$28 grayscale images belonging to 62 classes (upper and lower case letters and digits) already partitioned according to the writers. Thus, each writer is considered a client. The HAR dataset consists of sensor data (flattened into an 1152-valued vector) generated by users performing six possible actions (i.e., classes). To further study the impact of statistical heterogeneity on the performance, we follow prior works~\citep{data_free} and simulate Non-IID data on MNIST~\citep{mnist} dataset via Dirichlet sampling $Dir(\alpha)$, where a smaller value of $\alpha$ denotes greater heterogeneity (see Figure~\ref{fig:label_dist_alpha} in Appendix~\ref{sec:non_iid_data}). We employ the VGG9 and multilayer perceptron (MLP) for the image classification and activity recognition tasks, respectively, with model configurations (see Table~\ref{tab:model_config} in Appendix~\ref{sec:model_config}). We enable pruning for the last four convolutional layers in VGG9 and the first two hidden layers in MLP.

\subsection{System Implementation}

We implement \sysname\ and baselines with PyTorch (v1.8.0)~\citep{pytorch} on a server equipped with a single Nvidia RTX 3090 GPU. We experiment with a total of $K$ clients set to 160 and 320 for MNIST and EMNIST datasets and 30 for the HAR dataset. We randomly sample $\rho=10\%$ of participating clients that perform $E=5$ local epochs during each communication round with a total of 300 rounds for MNIST and EMNIST and 200 rounds for HAR. Weights and sign supermasks are initialized using Kaiming uniform \citep{kaiming_init} and uniform distribution $U(-1, 1)$ respectively. We perform one-shot pruning for 100 iterations in \sysname\ (according to \citep{synflow}) and one epoch for FedMask and Signed on the last four layers of the VGG9 and the first two hidden layers of the MLP with the pruning rate $p_r=0.8$ (80\% of the weights are kept). We employ the SGD optimizer with a learning rate $\eta=0.001$ for FedAvg, FedMask and Signed, $\eta=0.01$ for BNNAvg and $\eta=10$ for \sysname, and momentum $\mu=0.9$ for all algorithms chosen empirically. We repeat every experiment thrice with different seeds for reproducibility.

\subsection{Baselines}
We evaluate \sysname\ by comparing its performance against several baselines. We include \textbf{FedAvg}~\citep{mcmahan} to realize the performance of the model when trained at full capacity. \textbf{FedMask}~\citep{li2021fedmask} is the closest in spirit to our work and state-of-the-art when it comes to applying LTH for the federated setting with client-side pruning and learning binary supermasks. We also borrow their \textbf{BNNAvg} baseline which applies FedAvg to train binarized neural networks (BNN)~\citep{binary_net} with the weights and activations quantized by their signs. We also implement an extension of FedMask we call \textbf{Signed} where we replace binary supermask with sign supermask and change their binarizing function from sigmoid to tanh.

\section{Results}

\subsection{Training performance}
\label{subsec:train_perf}
We first compare the training performance by reporting the inference accuracies in Table \ref{tab:accuracies} between \sysname\ and the baselines. The inference accuracies were measured by taking a weighted average of the client's inference accuracies based on their local test data, which are weighted based on the number of test samples 
in their local dataset. While \sysname\ performs expectedly worse than the FedAvg which trains the full model and serves as the upper bound in training performance, \sysname\, in general, outperforms FedMask, Signed and BNNAvg across tasks. It is worthwhile to mention 
that the performance improvements are significant for HAR and MNIST datasets with lower heterogeneity at $\alpha = \{1, 10\}$ with inference accuracies higher by 24.1-40.6\% for \sysname\ compared to FedMask. \sysname\ performance gradually degrades for MNIST ($\alpha=0.1$) with higher heterogeneity and EMNIST with a large number of classes. This finding can be attributed to the fact that \sysname\ employs a shared global feature extractor among the baselines that utilize pruning. While it is challenging to learn generalized features among clients, the performance is still approximate 
to both FedMask and Signed which learn a more personalized feature extractor. Yet, \sysname\ scores higher than FedMask by 2.09\% and 19.62\% for EMNIST and MNIST($\alpha=0.1$). 
\begin{table}[h]
    \caption{Inferences accuracies of baselines and \sysname\ on different datasets. 
    }
    \small
    \centering
    \begin{tabular}{| c | c | c c c | c |}
        \hline
        Algorithm & \text{EMNIST} & \multicolumn{3}{c|}{MNIST} & \text{HAR} \\
        & \text{Non-IID} & $\alpha=0.1$ & $\alpha=1$ & $\alpha=10$ & \text{Non-IID} \\
        \hline
        FedAvg & 94.41 $\pm$ 0.06 & 98.81 $\pm$ 0.10  & 99.23 $\pm$ 0.05 & 99.38 $\pm$ 0.05 & 93.37 $\pm$ 0.22 \\
        BNNAvg & 29.09 $\pm$ 1.27 & 51.80 $\pm$ 0.45 & 54.39 $\pm$ 0.16 & 64.62 $\pm$ 0.15 & 66.50 $\pm$ 0.57 \\
        FedMask & 67.09 $\pm$ 0.23 & 57.73 $\pm$ 0.21 & 67.46 $\pm$ 0.40 & 84.41 $\pm$ 0.12 & 81.93 $\pm$ 0.55  \\
        Signed & 69.99 $\pm$ 0.76 & 59.45 $\pm$ 0.26 & 72.17 $\pm$ 0.19 & 86.69 $\pm$ 0.12 & 76.30 $\pm$ 1.04 \\
        \textbf{\sysname} & \textbf{69.18 $\pm$ 5.58} & \textbf{77.35 $\pm$ 0.81} & \textbf{94.97 $\pm$ 0.64} & \textbf{96.69 $\pm$ 0.56} & \textbf{90.60 $\pm$ 0.30} \\
        \hline
    \end{tabular}
    \label{tab:accuracies}
\end{table}

\begin{figure}[t!]
    \centering
    \subfloat[HAR]{
     \includegraphics[width=.3\textwidth]{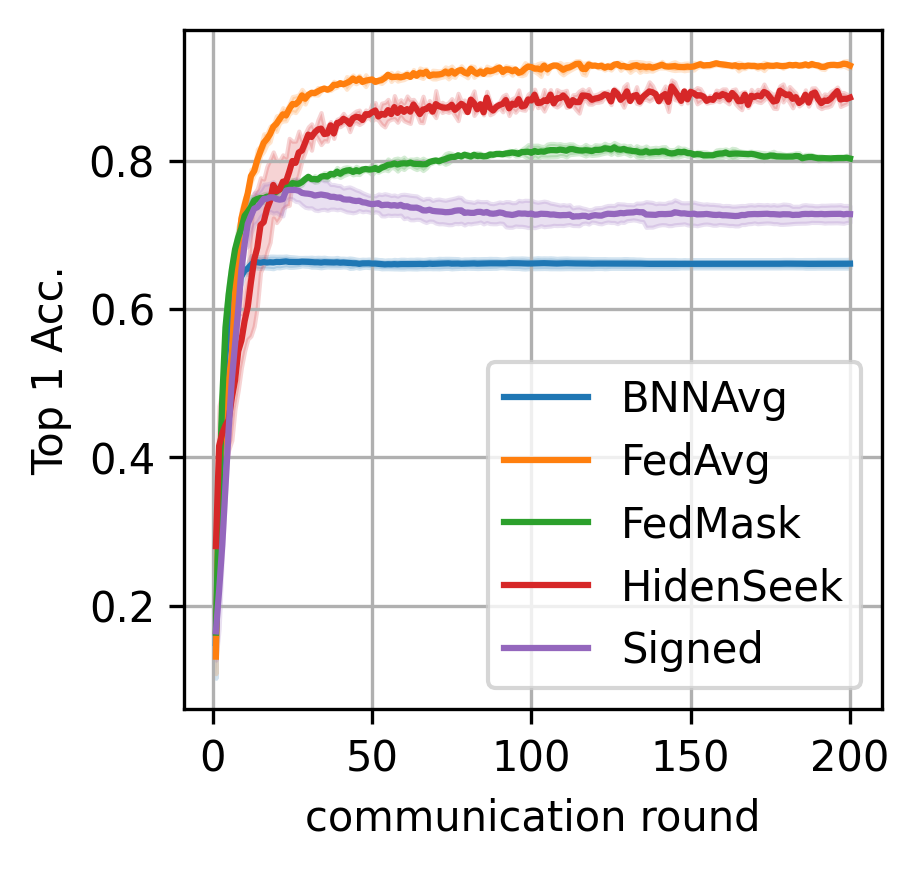} 
    }
    \subfloat[EMNIST]{
     \includegraphics[width=.3\textwidth]{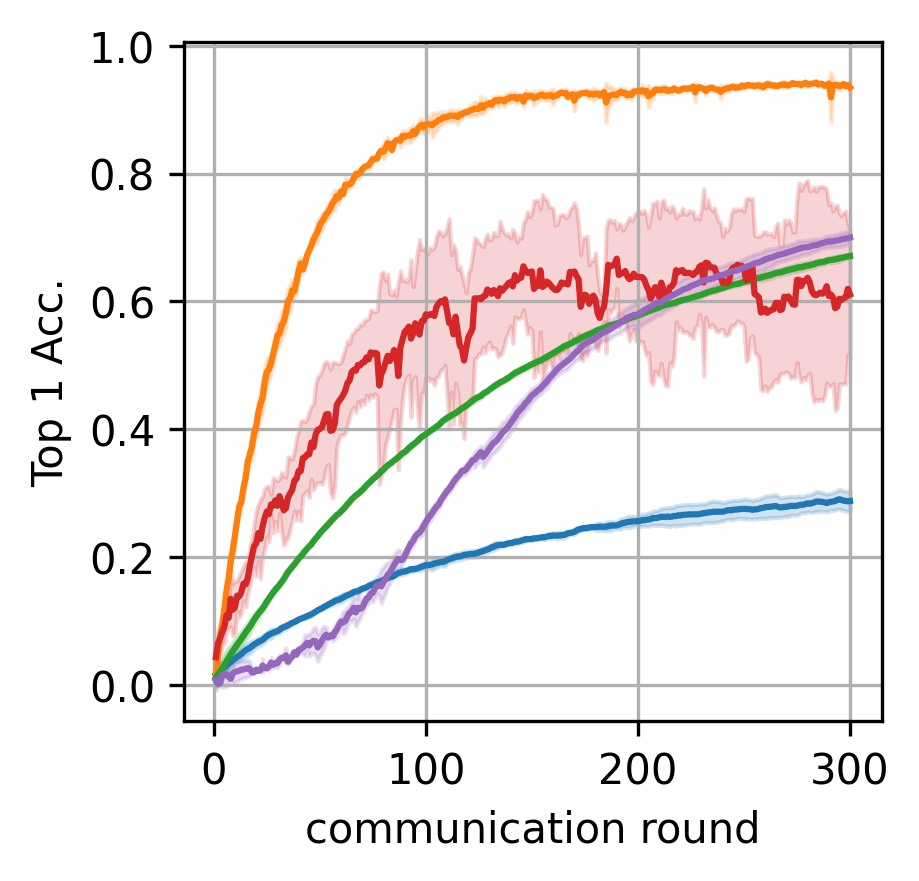}    
    }
    \caption{Performance on Non-IID EMNIST and HAR datasets.}
    \label{fig:convergence_har_emnist}
\end{figure}

\begin{figure}[t!]
    \centering
    \subfloat[$\alpha=0.1$]{
     \includegraphics[width=.3\textwidth]{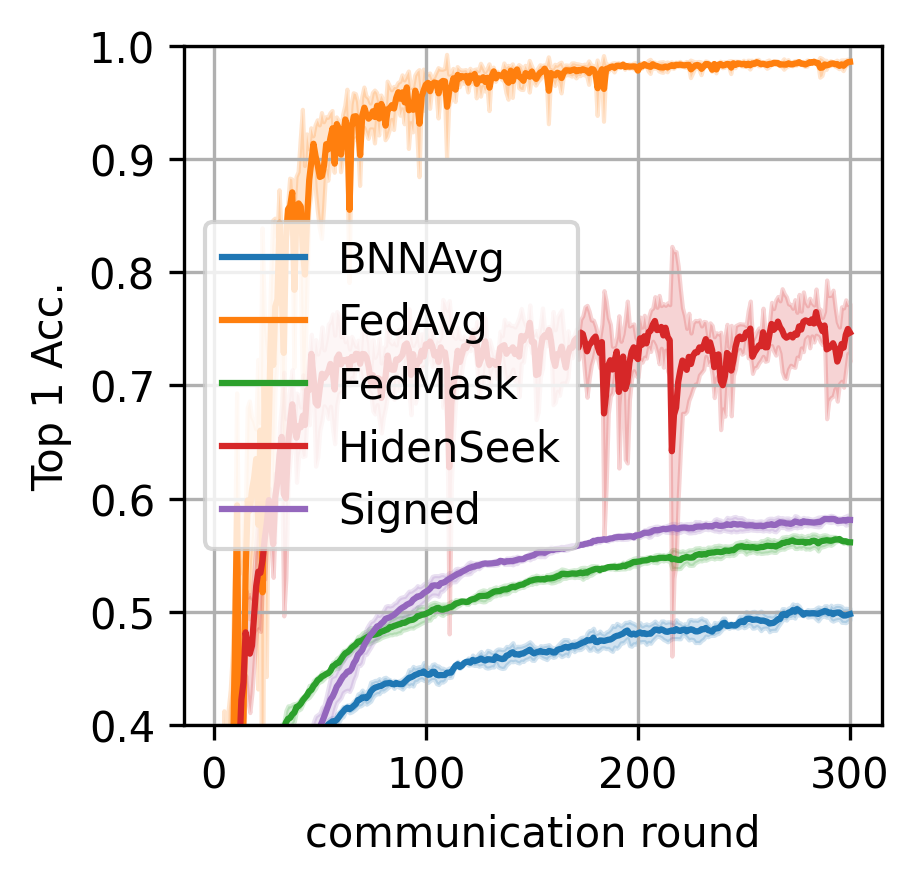}    
    }
    \subfloat[$\alpha=1$]{
     \includegraphics[width=.3\textwidth]{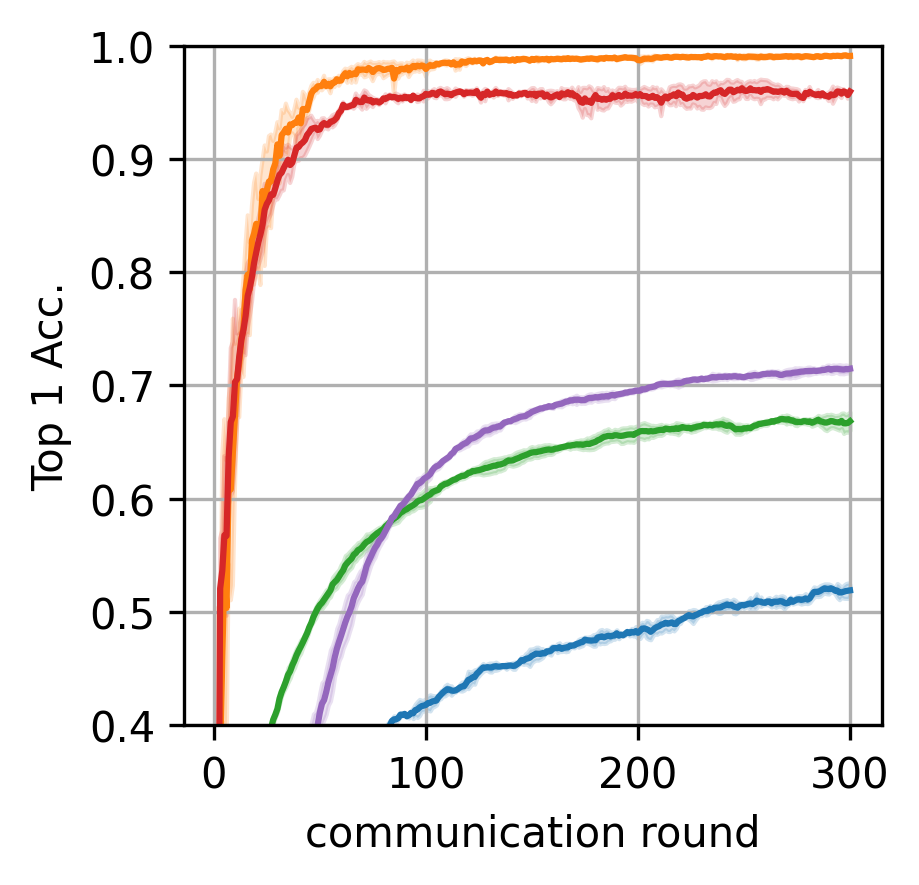}    
    }
    \subfloat[$\alpha=10$]{
     \includegraphics[width=.3\textwidth]{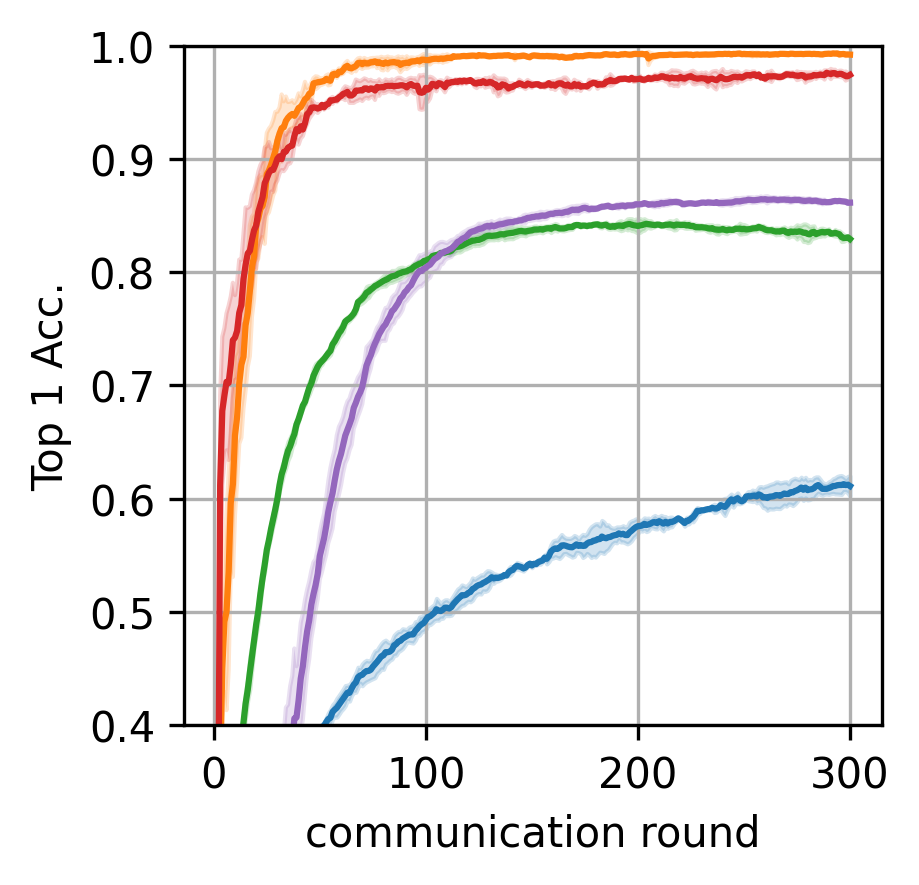}    
    }
    \caption{Performance on MNIST with different data heterogeneities.}
    \label{fig:convergence_mnist}
\end{figure}

We further compare the training performances by plotting the inference accuracies against the communication round. In the case of HAR dataset in Figure \ref{fig:convergence_har_emnist}(a) and MNIST dataset in Figure~\ref{fig:convergence_mnist}, \sysname\ converges faster than FedMask, Signed and BNNAvg. Whereas \sysname\ experiences higher volatility in training compared to the baselines in EMNIST dataset in Figure~\ref{fig:convergence_har_emnist}(b) and to some extent in MNIST dataset with higher heterogeneity at $\alpha = 0.1$ in Figure \ref{fig:convergence_mnist}(a). Both of which again point to the difficulty in training a shared global sign supermask under heterogeneous conditions.

\subsection{Communication Cost}
We then compare the communication cost for each client by measuring the upload and download sizes in MB for each client during each communication round as shown in Table~\ref{tab:comm_cost}. First, BNNAvg with binary parameters is four times smaller than FedAvg because 1 byte is the smallest element size to represent a parameter in PyTorch. FedMask and Signed have a lower upload cost compared to BNNAvg due to client-side pruning. \sysname\ further reduces the download cost 
thanks to server-side pruning. 
Furthermore, prune scores are more granular in \sysname\ compared to FedMask and Signed. This leads to smaller subnetworks since we drop all weights with scores equal to the threshold (see line~\ref{lst:threshold} in Algorithm \ref{alg:hidenseek}). Overall, \sysname\ demonstrates a reduction in communication cost compared to the second-best performance (FedMask) by 20.9-39.7\% times across all tasks. 

\begin{table}[h]
    \caption{Communication cost in MB for each client in a communication round 
    }
    \small
    \centering
    \begin{tabular}{|c | c c| c c| c c|}
        \hline
        Algorithm & \multicolumn{2}{c|}{EMNIST} & \multicolumn{2}{c|}{MNIST} & \multicolumn{2}{c|}{HAR} \\
        & \text{upload} & \text{download} & \text{upload} & \text{download} & \text{upload} & \text{download} \\
        \hline
        FedAvg & 4.53 & 4.53 & 4.33 & 4.33 & 1.44 & 1.44 \\
        BNNAvg & 1.07 & 1.07 & 1.07 & 1.07 & 0.36 & 0.36  \\
        FedMask & 0.74 & 1.07 & 0.70 & 1.07 & 0.27 & 0.36  \\
        Signed & 0.74 & 1.07 & 0.70 & 1.07 & 0.27 & 0.36  \\
        \textbf{\sysname} & \textbf{0.70} & \textbf{0.70} & \textbf{0.70} & \textbf{0.70} & \textbf{0.19} & \textbf{0.19} \\
        \hline
    \end{tabular}
    \label{tab:comm_cost}
\end{table}

\begin{table}[t]
    \caption{Training times in seconds of baselines and \sysname\ measure on Nvidia RTX 3090. 
    }
    \small
    \centering
    \begin{tabular}{|c | c | c | c |}
        \hline
        Algorithm & \text{EMNIST} & \text{MNIST} & \text{HAR}  \\
        \hline
        FedAvg & 475.61 $\pm$ 0.54 & 587.43 $\pm$ 0.50 & 49.31 $\pm$ 0.15 \\ 
        BNNAvg & 580.64 $\pm$ 0.42 & 693.76 $\pm$ 1.82 & 61.10 $\pm$ 0.42 \\ 
        FedMask & 1150.77 $\pm$ 5.53 & 1200.71 $\pm$ 3.86 & 83.52 $\pm$ 0.12 \\ 
        Signed & 979.90 $\pm$ 1.78 & 908.93 $\pm$ 3.46 & 76.34 $\pm$ 0.48 \\ 
        \textbf{\sysname} & \textbf{612.33 $\pm$ 2.35} & \textbf{705.16 $\pm$ 2.61} & \textbf{64.44 $\pm$ 0.72} \\ 
        \hline
    \end{tabular}
    \label{tab:train_times}
\end{table}

\begin{table}[t]
    \caption{Inference accuracies when varying the number of active clients per communication round.
    }
    \small
    \centering
    \begin{tabular}{|c | c | c | c |}
        \hline
        Algorithm & K=10 & K=20 & K=40 \\
        \hline
        FedAvg & 99.23 $\pm$ 0.05 & 99.19 $\pm$ 0.05 & 99.25 $\pm$ 0.07 \\
        BNNAvg & 54.39 $\pm$ 0.16 & 58.17 $\pm$ 0.16 & 61.78 $\pm$ 0.24 \\
        FedMask & 67.46 $\pm$ 0.40 & 69.51 $\pm$ 0.22 & 72.31 $\pm$ 0.05 \\
        Signed & 72.17 $\pm$ 0.19 & 73.02 $\pm$ 0.15 & 73.48 $\pm$ 0.01 \\
        \textbf{\sysname} & \textbf{94.97 $\pm$ 0.64} & \textbf{93.40 $\pm$ 0.61} & \textbf{91.05 $\pm$ 0.87} \\
        \hline
    \end{tabular}
    \label{tab:active_clients}
\end{table}
\begin{figure}[t]
    \centering
    \subfloat[$\alpha=0.1$]{
     \includegraphics[width=.3\textwidth]{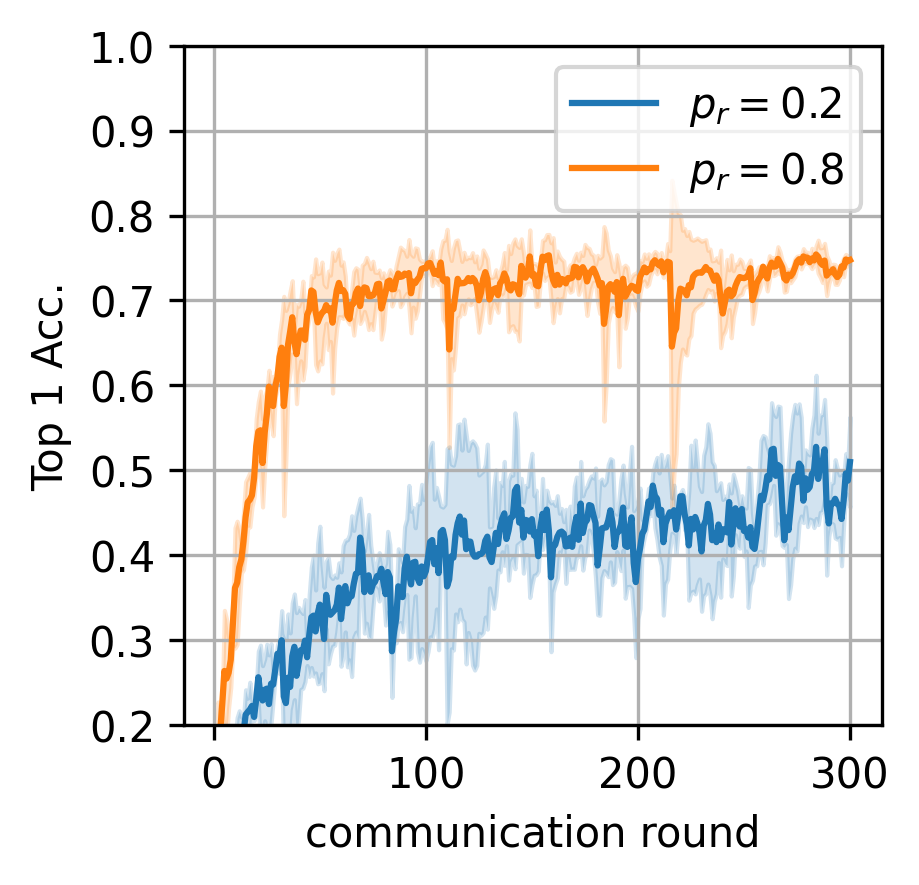}    
    }
    \subfloat[$\alpha=1$]{
     \includegraphics[width=.3\textwidth]{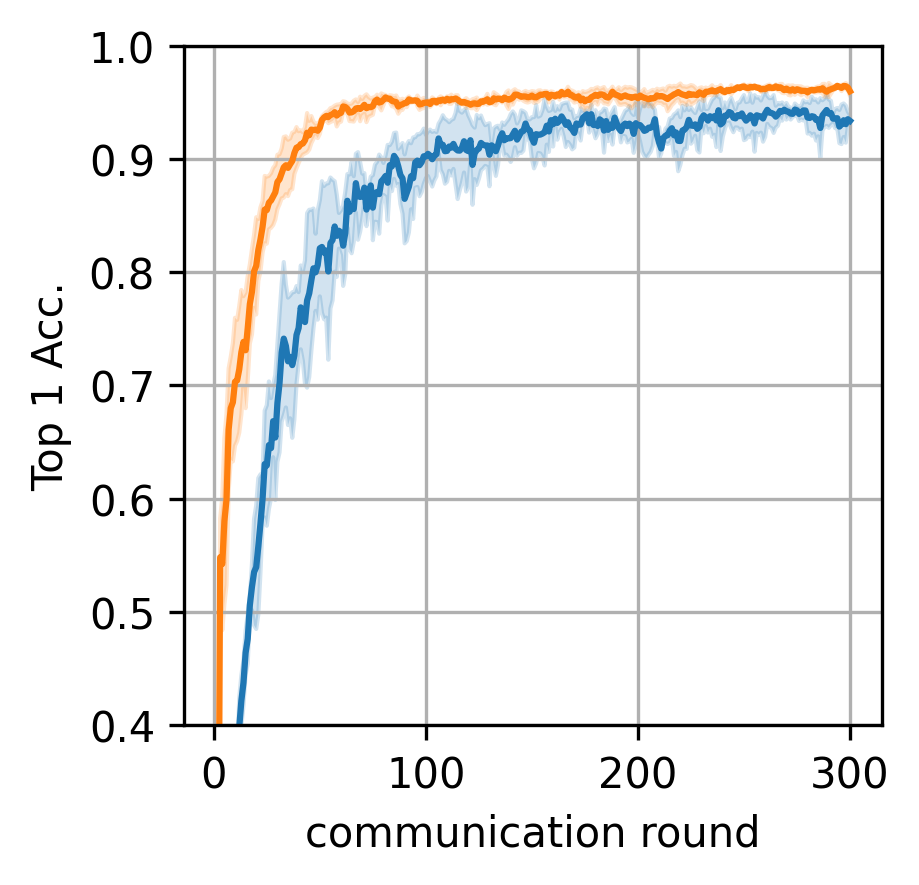}    
    }
    \subfloat[$\alpha=10$]{
     \includegraphics[width=.3\textwidth]{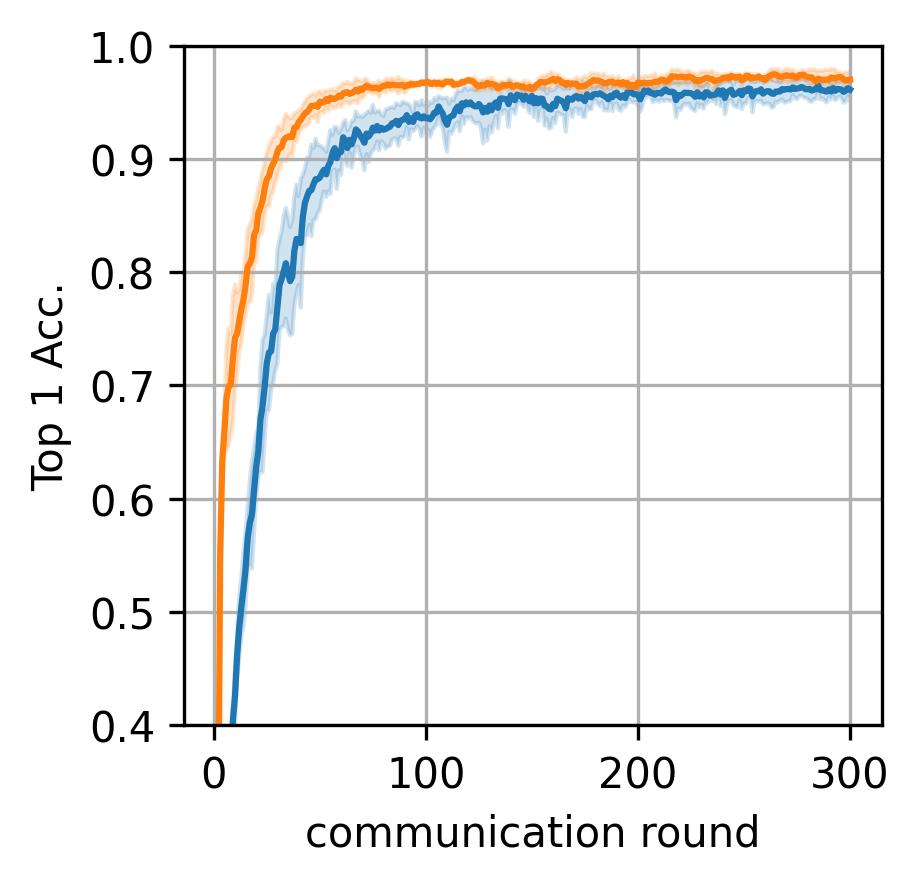}    
    }
    \caption{Performance on MNIST with different pruning rates.}
    \label{fig:prune_rate}
\end{figure}

\subsection{Computation Cost}
We report the computation cost by measuring the total training time on a single Nvidia RTX 3090 GPU  (Table~\ref{tab:train_times}). FedAvg is the fastest primarily because of that it does not utilize any latent weights, such as the masks in the case of Signed, FedMask and \sysname. Even BNNAvg is slower than FedAvg since the quantization must be performed at runtime and there is no built-in optimization in PyTorch when dealing with 1-bit parameters. Still, \sysname\ has lower training time compared to FedMask and Signed because these baselines require the client-side one-shot pruning each time a new client participates in the training. Additionally, FedMask is more computationally expensive as it employs sparsity regularization term. Overall, \sysname\ demonstrates a reduction in computation costs compared to FedMask by 22.8-46.8\% across all tasks.

\subsection{Active Clients}
We now evaluate the impact of the number of active clients per communication round on the training performance. Table \ref{tab:active_clients} demonstrates the inference accuracies among clients on the MNIST ($\alpha=1$) with different numbers of active clients $K = \{10, 20, 40\}$. While most baselines experience an improvement in accuracy with more 
active clients, \sysname\ experiences a minor drop of 3.92\% in performance when $K$ quadruples. Still, \sysname\ demonstrates better performance compared to BNNAvg, FedMask and Signed by a significant margin. This signifies the scalability and partly backs up the robustness of heterogeneity demonstrated in Figure~\ref{fig:convergence_mnist}. 

\subsection{Pruning Rate}
From the results discussed above, it is evident that the VGG9 model is overparametrized for the MNIST dataset proven by the high inference accuracies of FedAvg and \sysname. Hence, we tried a drastically more aggressive pruning rate of $p_r=0.2$ compared to previous experiments where $p_r=0.8$. As shown in Figure \ref{fig:prune_rate}, the drop in performance is very marginal in the less heterogeneous datasets $\alpha=\{1, 10\}$, while there is a significant drop in the more heterogeneous dataset $\alpha={0.1}$. This demonstrates that the computation and communication advantage of HnS over baselines is larger than the prior results without a noticeable accuracy drop in some cases when using a high prune rate.

\section{Related Work}
\label{sec:related}
\noindent \textbf{Statistical Heterogeneity.}
After the seminal work on federated learning \citep{mcmahan}, immediate advancements sought to tackle the problem of statistical heterogeneity in federated learning by adapting personalization schemes. PerFedAvg~\citep{fallah2020personalized} integrates a model-agnostic meta-learning approach into FedAvg for personalization. MOCHA~\citep{smith2017federated} introduces federated multi-task learning where each client is considered as a task. A plethora of works~\citep{ensemble_distill, attn_distill, data_free} has also applied knowledge distillation to learn a global surrogate model which teaches the clients' local models. 
\citep{li2021fedmask} performs personalization by allowing each client to learn a local binary supermask. In contrast, we employ personalization by globally sharing all hidden layers of the model while fine-tuning the final layer to the clients' local data. This allows \sysname\ to stably train the model by modifying weight magnitudes for a small subset of the weights while quantizing the updates transmitted for all the hidden layers. As such, \sysname\ reduces the communication cost while maintaining better ability in terms of learning data with varied heterogeneities as shown in Figure~\ref{fig:convergence_mnist}.


\vskip 0.1in \noindent \textbf{Communication and Computation Cost.}
Another significant issue in federated learning is the increased communication and computation cost on client devices when optimizing and transmitting the weights. FedProx~\citep{FedProx} alleviates this issue via allowing training preemption and partial updates, and FedPAQ~\citep{reisizadeh2020fedpaq} allows periodic averaging and quantizing model updates. Several works have also introduced variations of pruning and dropout \citep{heterofl, fjord, federated_dropout} for model compression. 
For example, FedMask applies LTH~\citep{LTH} by performing one-shot pruning at the client-side and learning a local binary supermask that is quantized during communication. However, the binary supermask learned is essentially unstructured pruning with no computational advantage and limits model capacity. We thus replace the binary supermask with a sign supermask for faster convergence and employ data-agnostic pruning at the server to reduce computational load on the client. 

    

\section{Conclusion \& Future Work}
\label{sec:conclusion}
In this work, we have introduced \sysname\ which applies the lottery ticket hypothesis under the federated setting by optimizing the signs of a synaptically salient subnetwork of the model. To further reduce computation load on the client, we perform one-shot pruning at initialization on the server-side using the data-agnostic approach and optimize a sign supermask that is quantized when transmitting model updates. Empirical results suggest that \sysname\ demonstrates better inference accuracy than the state-of-the-art in general while considerably reducing the communication cost and training time. 
Nevertheless, an imminent challenge faced is that the memory cost incurred by employing straight-through-estimators is substantial. Therefore, in the future, we will explore the efficacy brought by employing a binary optimizer~\citep{bop} that only modifies signs of weights without the need for latent parameters like the sign supermasks.

\section{Broader Impact}
In this work, we propose an algorithm in the field of federated learning which originated from the need to develop deep learning applications in the wake of recent advances in data protection regulations such as the GDPR~\citep{gdpr}. Furthermore, we explore an approach to reduce communication and computation costs on battery-powered mobile devices to reduce environmental impact. While our work demonstrates energy-saving implications from a theoretical standpoint, we hope future works will further delve into system optimization geared towards energy conservation.

\newpage
\bibliographystyle{plainnat}
\bibliography{ref}

\appendix


\newpage
\section{Algorithm}
\label{sec:algorithm}

Algorithm~\ref{alg:hidenseek} summarizes the training procedure of \sysname. Note that $\mathds{1}$ in line~\ref{lst:threshold} is a threshold function (as opposed to the identity matrix in Equation~\ref{eq:synaptic}). 

\begin{algorithm}[h]
    \DontPrintSemicolon
    \SetKwProg{proc}{Procedure}{:}{}
    \SetKwProg{func}{Function}{:}{}
    \proc{ServerRuns}{
        \KwIn{set of K clients $S \leftarrow \{C_1,C_2,\dots,C_K\}$ with data $(x_k,y_k)$ on $k$-th client device}
        randomly intialize DNN with weights $w^0$ \;
        initialize the global sign mask $s^0$ with signs of $w^0$ \;
        $w^0 \leftarrow $ \textit{ServerPruning}($w^0, s^0$) 
        \tcp*[l]{ one-shot pruning \Circled{1} in Figure \ref{fig:overview}}
        \For{each round $t = 1, 2, \dots, T$}{
            $c \leftarrow \max(K \times \rho, 1)$
            \tcp*[l]{select $c$ active clients from $K$ available clients with random sampling rate $\rho$}
            $S_t \leftarrow \{C_1,C_2,\dots,C_c\}$ \tcp*[l]{selected clients}
            \label{lst:active_clients}
            \For{$C_k \in S_t$ \textbf{in parallel}}{
                $s_k^{t+1} \leftarrow$ \textit{ClientUpdate($C_k, s^t \odot s^t_k$)}
                \tcp*[l]{\Circled{2} and \Circled{4} in Figure \ref{fig:overview}}
            }
            $s^{t+1} \leftarrow$ aggregate($\{s^{t+1}_1,\dots,s^{t+1}_c\}$) 
            \tcp*[l]{using Eq.~\eqref{eq:aggregation} (\Circled{5} in Figure \ref{fig:overview})}
        }
    }
    \func{ServerPruning($w, s$)}{
        \tcp*[l]{No. of iterations set to recommended value following \citep{synflow}}
        \For{each round $e = 1, 2, \dots 100$}{
            $S_{SF, i}^{[l]} \leftarrow \lVert w_{i}^{[l]} \odot \nabla_{s} R_{SF, i}^{[l]} \rVert_2$ \;
            $\tau \leftarrow p_r$ percentile score in $S_{SF}$ \;
            $m_{i}^{[l]} \leftarrow \mathds{1}(S_{SF, i}^{[l]} > \tau)$ \; \label{lst:threshold}
            $w_{i}^{[l]} \leftarrow w_{i}^{[l]} \odot m_{i}^{[l]}$
        }
        \Return $w$
    }
    \func{ClientUpdate($C_k,s^{t-1}_k$)}{
        \label{func:client_update}
        \tcp*[l]{( performs \Circled{3} in Figure~\ref{fig:overview})}
        $\hat{s}^t_k \leftarrow$ initialize real-valued sign mask from $s^{t-1}_k$ \;
        $\mathcal{B} \leftarrow$ split local data into batches \;
        \For{batch $(x_b,y_b) \in \mathcal{B}$}{
            $\hat{s}^{t}_k \leftarrow \hat{s}^{t}_k - \eta \nabla_{\hat{s}^{t}_k} \mathcal{L}[f(x_b;w^0 \odot \sign(\hat{s}^{t}_k)),y_b]$ 
            \tcp*[l]{$\eta$ is the learning rate}
        }
        $s^{t}_k \leftarrow \sign(\hat{s}^{t}_k)$
        \tcp*[l]{binarize the real-valued sign mask}
        \Return $s^{t}_k$
    }
\caption{\sysname}
\label{alg:hidenseek}
\end{algorithm}

\section{Simulating Non-IID Data}
\label{sec:non_iid_data}

\begin{figure}[t!]
    \centering
    \subfloat[$\alpha=0.1$]{
     \includegraphics[width=.3\textwidth]{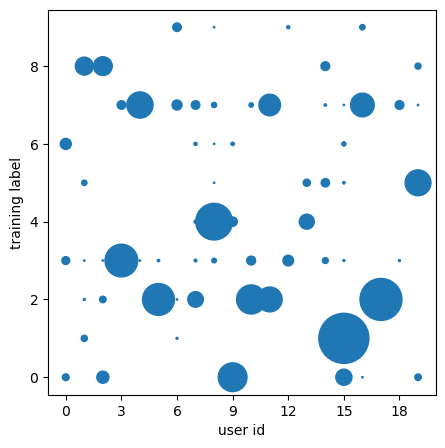}    
    }
    \subfloat[$\alpha=1$]{
     \includegraphics[width=.3\textwidth]{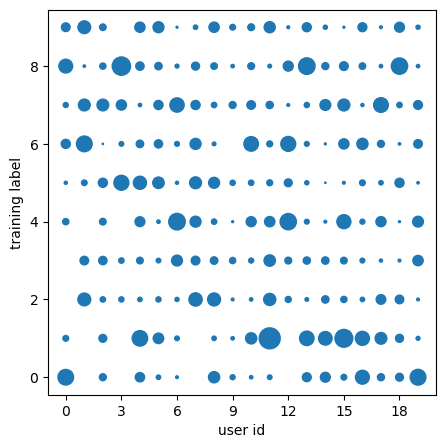}    
    }
    \subfloat[$\alpha=10$]{
     \includegraphics[width=.3\textwidth]{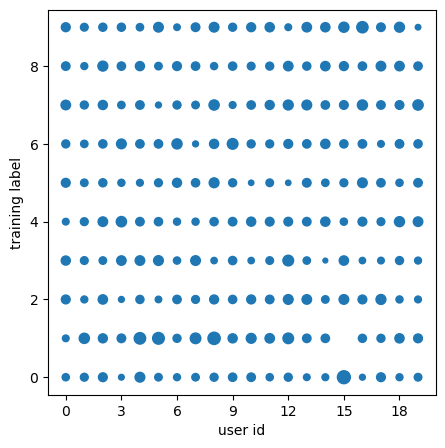}    
    }
    \caption{Label distribution among 20 clients for MNIST dataset with dirichlet sampling at different $\alpha$.}
    \label{fig:label_dist_alpha}
\end{figure}

Figure~\ref{fig:label_dist_alpha} depicts the effect of the parameter $\alpha$ on the label distribution among clients when employing Dirichlet sampling to partition MNIST dataset in Non-IID manner. 

\section{Model Configurations}
\label{sec:model_config}

The model configurations employed in this work are depicted in Table~\ref{tab:model_config}. For VGG9, each \texttt{ConvBlock(N)} is composed a convolutional layer with \texttt{N} channels with size 3, a BatchNorm layer followed by ReLU activation. Each MaxPool2d layer has a kernel size and stride length of 2. The number of nodes in the final fully connected layer are 62 or 10 depending on EMNIST or MNIST dataset.
\begin{table}[t!]
    \centering
    \begin{tabular}{c|c}
        \toprule
        VGG9                    & MLP \\
        \midrule
        \texttt{ConvBlock(32)}  & \texttt{Linear(300)} \\
        \texttt{MaxPool2d}      & \texttt{ReLU} \\
        \texttt{ConvBlock(64)}  & \texttt{Liner(100)} \\
        \texttt{MaxPool2d}      & \texttt{ReLU} \\
        \texttt{ConvBlock(128)} & \texttt{Linear(6)} \\
        \texttt{ConvBlock(128)} & \\
        \texttt{MaxPool2d}      & \\
        \texttt{ConvBlock(256)} & \\
        \texttt{ConvBlock(256)} & \\
        \texttt{MaxPool2d}      & \\
        \texttt{Flatten}        & \\
        \texttt{Linear(62 or 10)} & \\
        \bottomrule
    \end{tabular}
    \caption{Model configurations.}
    \label{tab:model_config}
\end{table}

\end{document}